\documentclass[10pt,twocolumn]{article}

\usepackage[margin=0.75in]{geometry}
\usepackage{times}
\usepackage{microtype}
\usepackage{amsmath}
\usepackage{amssymb}
\usepackage{booktabs}
\usepackage{graphicx}
\usepackage[hidelinks]{hyperref}
\hypersetup{
  pdftitle={Market-Alignment Risk in Pricing Agents: Trace Diagnostics and Trace-Prior RL under Hidden Competitor State},
  pdfauthor={Peiying Zhu, Sidi Chang}
}
\usepackage[numbers,sort&compress]{natbib}
\usepackage{tikz}
\usetikzlibrary{arrows.meta,positioning,calc}

\title{Market-Alignment Risk in Pricing Agents:\\
Trace Diagnostics and Trace-Prior RL under Hidden Competitor State}
\author{
Peiying Zhu\\
Blossom AI\\
San Francisco, USA
\and
Sidi Chang\\
Blossom AI Labs\\
Tokyo, Japan
}
\date{}

\newcommand{\RevPAR}{\mathrm{RevPAR}}
\newcommand{\Occ}{\mathrm{Occ}}
\newcommand{\ADR}{\mathrm{ADR}}
\newcommand{\KL}{D_{\mathrm{KL}}}
\newcommand{\JS}{D_{\mathrm{JS}}}
\newcommand{\piM}{\pi_M}
\newcommand{\piA}{\pi_\theta}

\begin{document}

\maketitle

\begin{abstract}
Outcome metrics can certify the wrong behavior. We study this failure in a
two-hotel revenue-management simulator where Hotel A trains an agent against a
fixed rule-based revenue-management competitor, Hotel B. A standard learning
agent can obtain near-reference revenue per available room (RevPAR) while
failing to learn market-like yield management: it sells too aggressively,
undercuts, or collapses to modal price buckets. We diagnose this as a
Goodhart-style failure under partial observability. Hotel A cannot observe the
competitor's remaining inventory, booking curve, or pricing rule, so the same
Hotel A-visible state maps to multiple plausible Hotel B prices. Deterministic
value-based RL and deterministic copying collapse this unresolved uncertainty
into shortcut behavior. We introduce a trace-level diagnostic protocol using
RevPAR, occupancy, ADR, full price-bucket distributions, L1/JS distances, and
seed-level confidence intervals. The verified repair is Trace-Prior RL: learn
a distributional market prior from lagged market traces, then train a stochastic
pricing policy with a RevPAR reward and a KL penalty to the learned prior. The
final policy matches Hotel B's RevPAR, occupancy, ADR, and price distribution
within seed-level uncertainty, while still optimizing Hotel A's own reward. We
argue that the contribution is not a new optimizer and not a hotel-pricing
leaderboard, but a reproducible failure-and-repair recipe for agentic systems
where scalar rewards are easy to game and the intended behavior is only visible
in traces. A key finding is that higher exact action accuracy can worsen
aggregate trace alignment when the target is distributional.
\end{abstract}
\vspace{0.2in}

\section{Introduction}

Outcome success can hide behavioral failure. In a two-hotel pricing simulator,
Hotel A can achieve competitive revenue per available room (RevPAR) against a
fixed revenue-management (RM) competitor while failing to learn market-like
yield management. It may sell too aggressively, undercut, or collapse into
common price buckets. The headline score says the agent is doing well; the
pricing trace says otherwise. This is a compact instance of Goodhart's Law:
when a proxy metric becomes the target, it can stop measuring the intended
behavior \citep{goodhart1975monetary,strathern1997improving}. In machine
learning, the same pattern appears as specification gaming, reward hacking, and
reward misspecification
\citep{krakovna2020specification,skalse2022defining,pan2022effects}.

The failure is not just poor optimization. Hotel A operates under partial
observability \citep{kaelbling1998planning}. It sees its own inventory, market condition, booking pace, and
lagged market prices, but it does not see Hotel B's remaining inventory,
booking curve, or pricing rule. As a result, the same Hotel A-visible context
can correspond to several valid Hotel B prices. The market target is therefore
not one correct action. It is a distribution.

This makes deterministic learning brittle. A DQN \citep{mnih2015human} trained only on revenue can
find shortcuts. A deterministic copy policy can achieve higher one-step
prediction accuracy by always choosing the most likely market bucket, but this
collapses hidden-state uncertainty and distorts the aggregate price trace. In
this setting, higher action accuracy can move in the opposite direction from
market alignment.

We propose Trace-Prior RL. First, Hotel A learns a distributional market prior
from observed traces. Then it optimizes its own RevPAR while staying close to
that learned prior through a KL penalty. The method does not give Hotel A Hotel
B's booking curve, inventory, or pricing formula. It only gives the agent a
disciplined way to preserve the uncertainty that remains after conditioning on
deployable context.

The same failure pattern is natural in agentic AI. Consider an LLM routing
agent that chooses which model or tool should handle a request. If reward
combines apparent success and cost, the router may overuse cheap routes even
when expert traces would send difficult tasks to stronger tools. The final
score can look good while the decision process has drifted away from expert
discipline. Our hotel simulator gives a controlled version of this problem:
the target behavior is not merely ``earn revenue'', but learn market-level
pricing discipline without being given the competitor's hidden state or rule.

Our contribution is a diagnosed failure and repair workflow, not a new
optimizer or hotel-pricing leaderboard. First, we show that scalar reward
evaluation is insufficient: near-reference RevPAR can hide non-market behavior.
Second, we induce trace-level diagnostics from the intended business logic:
occupancy, ADR, price-bucket distributions, L1/JS divergence, and seed-level
95\% confidence interval checks. Third, we document negative results from
plausible interventions: more exploration, longer Bellman horizons, reward
shaping, market forecasts as input, and deterministic copying. Fourth, we
identify the POMDP trigger empirically using an ambiguity diagnostic and an
oracle-\(q_B\) prediction ablation. Finally, we verify Trace-Prior RL: learn a
distributional market prior \(\piM(a\mid o)\) from traces, then train a
stochastic RL pricing policy with RevPAR reward and KL discipline to that
learned prior.

We do not claim that KL regularization to a reference policy is algorithmically
new. It is closely related to behavior-regularized RL and KL-control methods.
The contribution is the failure characterization and the role of the reference:
\(\piM\) is an empirical estimate of the market posterior predictive
distribution \(p(a_B\mid o)\), not Hotel A's previous behavior and not merely an
offline dataset support constraint.

\section{Environment And Goal}

\paragraph{Goal.}
Hotel A should learn market-level yield-management behavior \citep{talluri2004theory} from observable
market traces. The goal is not to beat Hotel B by undercutting. It is to match
market discipline: similar RevPAR, occupancy, ADR, and price distribution,
without observing Hotel B's booking curve, inventory, or pricing rule.
Hotel A is context-aware in the deployable sense: it conditions prices on own
inventory, market condition, booking pace, and lagged market prices. But it is
not fully competitor-state-aware, because the competitor's inventory and
pricing rule remain hidden. This partial context is exactly what makes the
target distributional.

\paragraph{Simulator.}
There are two hotels, \(i\in\{A,B\}\), capacity \(Q=100\), horizon \(H=30\),
and price grid
\[
P=[100,120,140,160,180,200,220].
\]
At each day \(t\), both hotels post price buckets \(a_{i,t}\) and prices
\[
p_{i,t}=P[a_{i,t}].
\]
The number of potential guests is
\[
M_t\sim \mathrm{Poisson}(\Lambda_t),\qquad
\Lambda_t=\Lambda_0 \exp(\eta_\Lambda m_t),
\]
with \(\Lambda_0=7\). Each guest has three choices: book Hotel A, book Hotel B,
or walk away. Hotel utility is
\[
v_{i,t}=\alpha-\rho p_{i,t}+\eta_m m_t .
\]
We use a nested-logit choice model \citep{mcfadden1974conditional} with nest parameter \(\mu=0.5\). The hotel
nest inclusive value is
\[
I_t=\mu \log\sum_{j\in\{A,B\}}\exp(v_{j,t}/\mu).
\]
The probability of entering the hotel nest is
\[
P(\mathrm{hotel})=\frac{\exp(I_t)}{1+\exp(I_t)},
\]
and the conditional hotel choice probability is
\[
P(i\mid \mathrm{hotel})=
\frac{\exp(v_{i,t}/\mu)}
{\sum_{j\in\{A,B\}}\exp(v_{j,t}/\mu)}.
\]
Thus the realized booking probabilities are
\[
\pi_{i,t}=P(\mathrm{hotel})P(i\mid \mathrm{hotel}),\quad
\pi_{0,t}=1-\pi_{A,t}-\pi_{B,t}.
\]
Demand is drawn as
\[
(D_{A,t},D_{B,t},D_{0,t})\sim
\mathrm{Multinomial}(M_t,[\pi_{A,t},\pi_{B,t},\pi_{0,t}]).
\]
Sales are capacity-capped:
\[
y_{i,t}=\min(q_{i,t},D_{i,t}),\qquad
q_{i,t+1}=q_{i,t}-y_{i,t}.
\]

\paragraph{Information asymmetry.}
Hotel B follows a deterministic Fixed RM rule
\[
a_{B,t}=\mathrm{RM}(t,q_{B,t},m_t).
\]
In implementation, this is a hand-tuned bid-price and booking-pace style rule:
it raises prices as inventory tightens, as the stay date approaches, and as
market conditions strengthen.
Hotel A observes its own state and, in the final deployable setup, the last
three observed Hotel B prices. It does not observe \(q_{B,t}\), Hotel B's
booking curve, or Hotel B's formula. This matters because undercutting changes
the market dynamics:
\[
p_{A,t}<p_{B,t}
\Rightarrow \pi_{A,t}\uparrow,\ \pi_{B,t}\downarrow
\Rightarrow y_{B,t}\downarrow
\Rightarrow q_{B,t+1}\ \mathrm{changes}.
\]
The shortcut is therefore not only undesirable business behavior. It also
distorts the future market trace from which Hotel A is trying to learn.
Equivalently, Hotel A and Hotel B are two competing service pipelines. Each
posted price controls how much demand enters each pipeline, so Hotel A's
learning actions alter both its own inventory process and the competitor's
future state.

\paragraph{Posterior-predictive market target.}
Let \(o_t\) be Hotel A's deployable observation. Since \(q_B\) is hidden, the
market-alignment target is not a point label but
\[
p(a_B\mid o_t)=
\sum_{q_B}
\mathbf{1}\{\mathrm{RM}(t,q_B,m_t)=a_B\}
p(q_B\mid o_t).
\]
The learned market prior \(\piM(a\mid o_t)\) is an empirical estimate of this
posterior predictive distribution from traces. This is the theoretical reason
for distributional imitation: if \(q_B\) remains uncertain after conditioning
on \(o_t\), then multiple Hotel B prices are valid for the same Hotel A-visible
state.

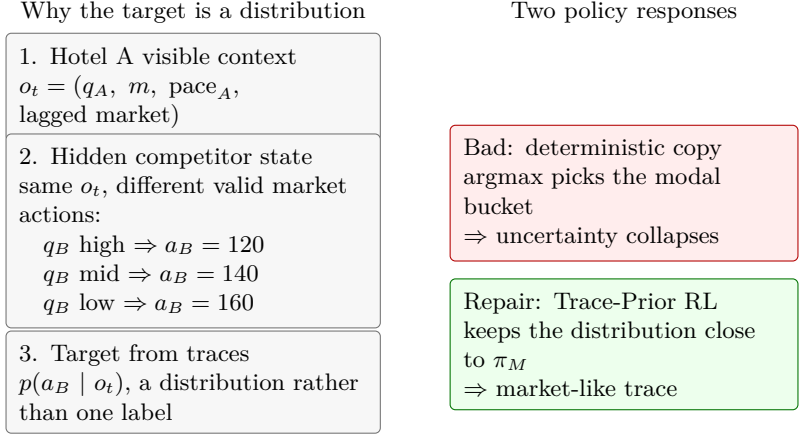
\begin{figure*}[t]
\centering
\small
\begin{tikzpicture}[
  box/.style={draw, rounded corners=2pt, align=left, inner sep=5pt},
  visible/.style={box, text width=4.6cm, minimum height=1.12cm, draw=black!55, fill=black!3},
  hidden/.style={box, text width=4.6cm, minimum height=1.45cm, draw=black!55, fill=black!3},
  target/.style={box, text width=4.6cm, minimum height=1.05cm, draw=black!55, fill=black!3},
  outcome/.style={box, text width=4.25cm, minimum height=1.05cm},
  bad/.style={draw=red!70!black, fill=red!7},
  good/.style={draw=green!45!black, fill=green!7},
  title/.style={font=\bfseries}
]
\node[align=center, font=\bfseries] at (0,3.15) {Why the target is a distribution};
\node[align=center, font=\bfseries] at (5.7,3.15) {Two policy responses};
\node[visible] (obs) at (0,2.15) {
\textbf{1. Hotel A visible context}\\
\(o_t=(q_A,\ m,\ \mathrm{pace}_A,\)\\
\(\mathrm{lagged\ market})\)
};
\node[hidden] (hid) at (0,0.25) {
\textbf{2. Hidden competitor state}\\
same \(o_t\), different valid market actions:\\
\quad \(q_B\) high \(\Rightarrow a_B=120\)\\
\quad \(q_B\) mid \(\Rightarrow a_B=140\)\\
\quad \(q_B\) low \(\Rightarrow a_B=160\)
};
\node[target] (dist) at (0,-1.75) {
\textbf{3. Target from traces}\\
\(p(a_B\mid o_t)\), a distribution rather than one label
};
\node[outcome,bad] (bad) at (5.7,0.75) {
\textbf{Bad: deterministic copy}\\
argmax picks the modal bucket\\
\(\Rightarrow\) uncertainty collapses
};
\node[outcome,good] (good) at (5.7,-1.25) {
\textbf{Repair: Trace-Prior RL}\\
keeps the distribution close to \(\pi_M\)\\
\(\Rightarrow\) market-like trace
};
\end{tikzpicture}
\caption{Failure mechanism. Hidden competitor inventory means the same Hotel
A-visible observation can imply multiple valid Hotel B prices. Deterministic
argmax copying collapses this posterior uncertainty; Trace-Prior RL preserves
it through a learned market distribution.}
\label{fig:mechanism}
\end{figure*}

\section{Trace-Level Diagnostics}

The business metrics are
\[
\RevPAR_i=\frac{1}{Q}\sum_{t=0}^{H-1}p_{i,t}y_{i,t},
\]
\[
\Occ_i=\frac{1}{Q}\sum_{t=0}^{H-1}y_{i,t},
\]
\[
\ADR_i=
\frac{\sum_{t=0}^{H-1}p_{i,t}y_{i,t}}
{\sum_{t=0}^{H-1}y_{i,t}}.
\]
RevPAR alone catches whether revenue is competitive, but not how it was earned.
High occupancy can mean the policy sold too aggressively. Low ADR can mean the
policy lacked rate discipline. We therefore evaluate the full trace.

Let \(d_A\) and \(d_B\) be empirical distributions over the seven price buckets.
We use
\[
L1(d_A,d_B)=\sum_{k=1}^{7}|d_A(k)-d_B(k)|
\]
and Jensen-Shannon divergence
\begin{align*}
\JS(d_A,d_B)
&=\frac{1}{2}\KL(d_A\mid\mid m)
  +\frac{1}{2}\KL(d_B\mid\mid m),\\
m&=\frac{d_A+d_B}{2}.
\end{align*}
We also check whether Hotel B's aggregate RevPAR, occupancy, and ADR fall inside
Hotel A's seed-level 95\% confidence intervals. These diagnostics are induced
from the failure itself: a policy can pass on RevPAR while failing through
excess occupancy, low ADR, or distorted price-bucket use.

\section{Negative Results: Reward Optimization Is Not Enough}

\paragraph{Functions first.}
The original DQN reward was gross revenue
\[
r^{\mathrm{gross}}_t=p_{A,t}y_{A,t}.
\]
We also used RevPAR units:
\[
r^{\mathrm{revpar}}_t=\frac{p_{A,t}y_{A,t}}{Q}.
\]
The \(n\)-step DQN target is
\[
G^{(n)}_t=
\sum_{k=0}^{n-1}\gamma^k r_{t+k}
+\gamma^n\max_a Q_{\theta^-}(o_{t+n},a),
\]
with \(\gamma=1.0\). For the CMDP-style undercut penalty, define relative
bucket gap
\[
z_t=a_{A,t}-a_{\mathrm{ref},t},
\]
and sold-unit undercut cost
\[
c_t=\max(0,-z_t)^2 y_{A,t}.
\]
The shaped reward is
\[
r^{\mathrm{CMDP}}_t=\frac{p_{A,t}y_{A,t}}{Q}-\lambda_t c_t,
\]
with dual update
\[
\lambda_{k+1}=\max(0,\lambda_k+\eta_\lambda(\bar c_k-c_{\mathrm{target}})).
\]
When a market forecast was used as input, the forecast head was trained by
\[
\hat{\pi}_{B,t}=f_\phi(o_t),\qquad
\mathcal{L}_{\mathrm{forecast}}=-\log \hat{\pi}_{B,t}(a_{B,t}),
\]
but the DQN still chose actions by reward optimization.

\begin{table*}[t]
\centering
\small
\caption{Curated negative-result path. Each row rules out a plausible
explanation for the failure.}
\label{tab:negative}
\resizebox{\textwidth}{!}{
\begin{tabular}{lllll}
\toprule
Hypothesis & Intervention & Representative result & What failed & Lesson \\
\midrule
Needs exploration & Higher epsilon, 5k episodes, 3 seeds
& RevPAR \(\approx 104\), occ. \(\approx95\%\), ADR \(\approx109\)
& Still myopic low-rate behavior & Not just exploration \\
Needs longer credit assignment & \(n=5\) DQN
& Near-RM RevPAR in pilot, but pricing close to fixed mixture
& No state-dependent yield behavior & Not just horizon \\
Reward needs constraint & RevPAR-unit CMDP undercut cost
& RevPAR 107.730 vs 107.117, ADR gap \(-0.60\), L1 0.0459
& Good business score, distorted tail & Penalty helps but is brittle \\
Needs to see the market & Forecast-as-input DQN
& RevPAR gap \(+0.547\), occ. gap \(+0.0105\), ADR gap \(-1.18\)
& Forecast seen, action still shortcuts & Input is not enough \\
Copying is enough & Argmax market copy
& Accuracy 78.14\%, RevPAR gap \(+1.841\), occ. gap \(+0.0162\)
& Over-concentrates modal buckets & Deterministic copy collapses uncertainty \\
\bottomrule
\end{tabular}}
\end{table*}

Table~\ref{tab:negative} summarizes the path. The first smoke test collapsed
to the lowest price bucket: RevPAR 98.3, occupancy 98\%, and 100\% of actions
at price 100. More exploration prevented total collapse, but the agent still
sold too cheaply. Increasing the Bellman horizon did not by itself create
yield-management behavior. Reward shaping and CMDP penalties improved some
business metrics, but created fragile tradeoffs between ADR, occupancy, and
RevPAR. Giving the DQN a market forecast also did not fix the action rule:
the model could see market information and still convert it into an
occupancy-heavy shortcut.

This is not a claim that value-based RL is generally bad. The narrower claim is
that greedy reward optimization is poorly matched to this market-learning
objective under hidden competitor state. If several competitor prices are
plausible under the same observation, a deterministic action can turn
uncertainty into a systematic shortcut.

\section{POMDP Trigger And Epistemic Collapse}

\paragraph{Function first.}
The deterministic copy baseline has no RL reward. It is supervised market
prediction followed by an argmax decision:
\[
\hat{\pi}_{B,t}=f_\phi(o_t),\qquad
\mathcal{L}_{\mathrm{copy}}=-\log \hat{\pi}_{B,t}(a_{B,t}),
\]
\[
a_{A,t}=\arg\max_a \hat{\pi}_{B,t}(a).
\]
In plain language: predict Hotel B's current price distribution, then collapse
that distribution to the single most likely bucket.

\paragraph{Empirical ambiguity.}
We replayed the final Trace-Prior RL policy for five seeds and 2,000 evaluation
episodes per seed, then grouped states into coarse Hotel A-visible cells:
3-day time band, own-inventory bucket, market-condition quantile, own previous
sales bin, and exact last three Hotel B price buckets. If Hotel A's observation
fully determined Hotel B's action, these cells would be nearly single-action.

\begin{table}[t]
\centering
\small
\caption{Observable-state ambiguity under final-policy rollouts.}
\label{tab:ambiguity}
\begin{tabular}{lr}
\toprule
Metric & Value \\
\midrule
Visited steps & 300,000 \\
Eligible cells & 1,951 \\
Eligible step share & 60.78\% \\
Cells with \(\ge2\) B actions & 95.08\% \\
Cells with \(\ge2\) B actions each \(\ge5\%\) & 85.29\% \\
Eligible steps in substantive ambiguous cells & 76.57\% \\
Weighted normalized within-cell entropy & 0.2800 \\
Weighted modal B-action share & 76.75\% \\
\bottomrule
\end{tabular}
\end{table}

Table~\ref{tab:ambiguity} shows that ambiguity is not rare. Under the same or
nearby Hotel A-visible state, Hotel B often takes multiple prices. This
supports the posterior-predictive framing in Section 2.

\paragraph{Oracle-\(q_B\) ablation.}
We then trained two supervised Hotel B price predictors on the same rollout
data. The observable predictor sees only deployable \(ca\_lag3\) features. The
oracle predictor sees the same features plus Hotel B remaining inventory
\(q_B/Q\). No RL training is performed in this diagnostic.

\begin{table}[t]
\centering
\small
\caption{Revealing hidden \(q_B\) sharply reduces market-prediction
uncertainty. This is an explanatory diagnostic only: oracle \(q_B\) is not
available to Hotel A and is not used by Trace-Prior RL.}
\label{tab:oracle}
\resizebox{\columnwidth}{!}{
\begin{tabular}{lrrrrr}
\toprule
Predictor & NLL & Acc. & Brier & True prob. & Norm. ent. \\
\midrule
Observable \(ca\_lag3\) & 0.5359 & 76.91 & 0.3160 & 0.6865 & 0.2748 \\
Oracle \(ca\_lag3+q_B\) & 0.1557 & 95.47 & 0.0827 & 0.8867 & 0.1210 \\
\bottomrule
\end{tabular}}
\end{table}

Table~\ref{tab:oracle} gives causal evidence for the hidden-state explanation:
when \(q_B\) is revealed, prediction becomes much sharper. This does not prove
that pure revenue RL would behave correctly with oracle state, but it shows
that hidden \(q_B\) is a major source of market-label uncertainty.

\paragraph{Epistemic collapse.}
If \(H(a_B\mid o_t)>0\) because \(q_B\) is not identified by \(o_t\), then any
deterministic rule \(a=g(o_t)\) cannot in general match \(p(a_B\mid o_t)\). It
replaces posterior uncertainty with a point action. We call this epistemic
collapse. In our deterministic copy baseline, argmax copying had higher exact
one-step accuracy but worse aggregate alignment than probability matching
(Table~\ref{tab:copy}).

\begin{table}[t]
\centering
\small
\caption{Argmax improves exact action accuracy but worsens aggregate market
alignment. Five seeds, 10,000 evaluation episodes per seed.}
\label{tab:copy}
\resizebox{\columnwidth}{!}{
\begin{tabular}{lrrrrrr}
\toprule
Decision rule & Acc. & RevPAR gap & Occ. gap & ADR gap & L1 & JS \\
\midrule
Argmax copy & 78.14 & +1.841 & +0.0162 & -0.56 & 0.0323 & 0.0002 \\
Prob. match, \(T=0.95\) & 69.50 & +0.221 & +0.0036 & -0.36 & 0.0183 & 0.0001 \\
\bottomrule
\end{tabular}}
\end{table}

The central tradeoff is that exact action accuracy can improve while trace
alignment gets worse. Under hidden state, the market target is distributional.
Sampling from the predicted distribution preserves uncertainty that argmax
destroys.

\section{Verified Repair: Trace-Prior RL}

\paragraph{Function first.}
Trace-Prior RL has two layers. First, learn the market prior:
\[
\hat{\pi}_{B,t}=f_\phi(o_t),\qquad
\mathcal{L}_{\mathrm{prior}}=-\log \hat{\pi}_{B,t}(a_{B,t}).
\]
This prior is frozen and used as
\[
\piM(a\mid o_t)=\hat{\pi}_{B,t}(a).
\]
Second, train a stochastic RL policy
\[
\piA(a\mid o_t)
\]
with per-step reward
\[
r^{\mathrm{TPRL}}_t=
\frac{p_{A,t}y_{A,t}}{Q}
-\beta
\KL\!\left(\piA(\cdot\mid o_t)\mid\mid\piM(\cdot\mid o_t)\right).
\]
Equivalently, maximize
\begin{align*}
\mathbb{E}\Bigg[
\sum_{t=0}^{H-1}
\bigg(
\frac{p_{A,t}y_{A,t}}{Q}
-\beta
\KL(&\piA(\cdot\mid o_t)\\
&\mid\mid\piM(\cdot\mid o_t))
\bigg)
\Bigg].
\end{align*}

In human terms, Hotel A still learns from its own sales and revenue, but it
pays a cost if its whole pricing distribution drifts away from the learned
market rule. This is not a hand-coded undercut penalty. The discipline comes
from the market trace.

This links directly to the POMDP argument. The learned prior \(\piM(a\mid o_t)\)
is an empirical estimate of the posterior predictive target \(p(a_B\mid o_t)\).
As \(\beta\) increases, \(\piA\) shrinks toward that market-alignment target;
as \(\beta\) decreases, the policy moves toward unconstrained revenue
optimization.

\begin{table}[t]
\centering
\small
\caption{Final Trace-Prior RL result. Five seeds, 2,000 train episodes per
seed, 10,000 evaluation episodes per seed, \(\beta=30\), entropy coefficient 0.}
\label{tab:final}
\resizebox{\columnwidth}{!}{
\begin{tabular}{lrrrr}
\toprule
Metric & Hotel A & Hotel B & Gap & B in A 95\% CI? \\
\midrule
RevPAR & 108.178 & 108.066 & +0.112 & yes \\
Occupancy & 0.7709 & 0.7680 & +0.0029 & yes \\
ADR & 140.33 & 140.71 & -0.38 & yes \\
L1 & \multicolumn{2}{c}{} & 0.0196 & -- \\
JS & \multicolumn{2}{c}{} & 0.0001 & -- \\
\bottomrule
\end{tabular}}
\end{table}

\begin{table}[t]
\centering
\small
\caption{Final price-bucket distribution.}
\label{tab:dist}
\resizebox{\columnwidth}{!}{
\begin{tabular}{lrrrrrrr}
\toprule
Policy & 100 & 120 & 140 & 160 & 180 & 200 & 220 \\
\midrule
Hotel A & 4.64 & 43.44 & 22.45 & 15.12 & 10.72 & 3.62 & 0.00 \\
Hotel B & 4.88 & 42.58 & 22.51 & 15.17 & 11.35 & 3.50 & 0.01 \\
\bottomrule
\end{tabular}}
\end{table}

Tables~\ref{tab:final} and \ref{tab:dist} show the selected result. Hotel B's
RevPAR, occupancy, and ADR all fall inside Hotel A's seed-level 95\%
confidence intervals, and the full price distribution is close.

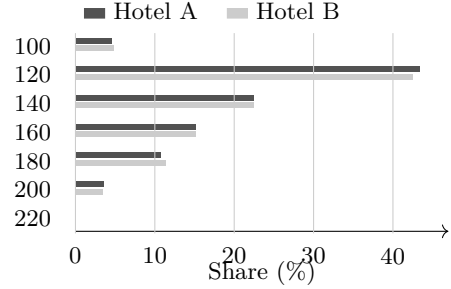
\begin{figure}[t]
\centering
\small
\begin{tikzpicture}[x=0.105cm,y=0.38cm]
  \foreach \y/\price/\a/\b in {
    6/100/4.64/4.88,
    5/120/43.44/42.58,
    4/140/22.45/22.51,
    3/160/15.12/15.17,
    2/180/10.72/11.35,
    1/200/3.62/3.50,
    0/220/0.00/0.01
  } {
    \node[anchor=east] at (-2,\y) {\price};
    \fill[black!68] (0,\y+0.10) rectangle (\a,\y+0.28);
    \fill[black!20] (0,\y-0.16) rectangle (\b,\y+0.02);
  }
  \draw[->] (0,-0.45) -- (47,-0.45);
  \foreach \x in {0,10,20,30,40} {
    \draw[black!20] (\x,-0.52) -- (\x,6.45);
    \node[anchor=north] at (\x,-0.72) {\x};
  }
  \node[anchor=south west] at (0,6.65) {\textcolor{black!68}{\rule{0.28cm}{0.14cm}} Hotel A};
  \node[anchor=south west] at (18,6.65) {\textcolor{black!20}{\rule{0.28cm}{0.14cm}} Hotel B};
  \node[anchor=north] at (23.5,-1.18) {Share (\%)};
\end{tikzpicture}
\caption{Final price-bucket distribution under Trace-Prior RL.}
\label{fig:pricebars}
\end{figure}

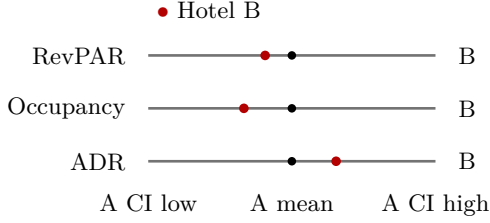
\begin{figure}[t]
\centering
\small
\begin{tikzpicture}[x=3.8cm,y=0.7cm]
  \foreach \y/\name/\bpos/\aval/\bval in {
    2/RevPAR/0.407/108.178/108.066,
    1/Occupancy/0.333/0.7709/0.7680,
    0/ADR/0.654/140.33/140.71
  } {
    \node[anchor=east] at (-0.05,\y) {\name};
    \draw[black!55,line width=1pt] (0,\y) -- (1,\y);
    \fill[black] (0.5,\y) circle (1.6pt);
    \fill[red!70!black] (\bpos,\y) circle (1.8pt);
    \node[anchor=west] at (1.05,\y) {B};
  }
  \node[anchor=north] at (0,-0.45) {A CI low};
  \node[anchor=north] at (0.5,-0.45) {A mean};
  \node[anchor=north] at (1,-0.45) {A CI high};
  \node[anchor=south west] at (0,2.55) {\textcolor{red!70!black}{\(\bullet\)} Hotel B};
\end{tikzpicture}
\caption{Seed-level CI check for RevPAR, occupancy, and ADR.}
\label{fig:ci}
\end{figure}

\paragraph{Why the KL term matters.}
The KL term regularizes the whole action distribution, not only the sampled
action. If the policy begins to overuse low prices to grab occupancy, the KL
cost rises even when a particular sampled action earns revenue. This is why the
method differs from a simple action-level bonus.

\begin{table}[t]
\centering
\small
\caption{KL strength sensitivity. No or weak KL is less stable and less
aligned; strong KL restores market discipline.}
\label{tab:beta}
\resizebox{\columnwidth}{!}{
\begin{tabular}{rrrrrrr}
\toprule
\(\beta\) & RevPAR gap & Occ. gap & ADR gap & L1 & JS & Read \\
\midrule
0 & -3.706 & -0.0296 & +1.19 & 0.1158 & 0.0021 & no discipline \\
1 & -0.907 & -0.0063 & +0.09 & 0.0263 & 0.0002 & partial repair \\
30 & +0.029 & +0.0024 & -0.39 & 0.0206 & 0.0001 & aligned \\
\bottomrule
\end{tabular}}
\end{table}

Table~\ref{tab:beta} should be read as a mechanism check, not as a universal
claim that \(\beta=30\) transfers across domains. The KL coefficient is tied to
the RevPAR reward scale. In our environment, one room sold at price 140
contributes \(1.4\) to per-step RevPAR reward, while the post-training KL term
is typically small. Values near 10--30 make the prior strong enough to prevent
policy drift without overwhelming revenue learning.
The \(\beta=0\) row is the business warning: reward-only RL does not become a
reliable market learner, even when its scalar metrics look plausible in some
seeds. It can move occupancy, ADR, and competitor inventory paths in ways that
are not stable market discipline.

We also tested the simpler reviewer baseline: replace full-distribution KL with
an action-level prior bonus,
\[
r_t=\frac{p_{A,t}y_{A,t}}{Q}+\beta\log\pi_M(a_t\mid o_t).
\]
This rewards sampled actions that the market prior likes, but it does not
directly constrain the rest of \(\pi_\theta(\cdot\mid o_t)\). Even the best
action-bonus setting in a small grid remained much less aligned than full KL
(Table~\ref{tab:bcaux}).

\begin{table}[t]
\centering
\small
\caption{Full-distribution KL versus sampled-action prior bonus. Five seeds,
2,000 train episodes and 5,000 evaluation episodes per seed.}
\label{tab:bcaux}
\resizebox{\columnwidth}{!}{
\begin{tabular}{lrrrr}
\toprule
Method & RevPAR gap & ADR gap & L1 & Policy-prior KL \\
\midrule
Full KL, \(\beta=30\) & +0.029 & -0.39 & 0.0206 & 0.0105 \\
Action bonus, best grid & -2.336 & +0.53 & 0.0729 & 0.2181 \\
\bottomrule
\end{tabular}}
\end{table}

\paragraph{Calibration and entropy.}
The frozen market prior is reasonably calibrated on final-policy evaluation
rollouts: NLL 0.5382, Brier 0.3173, ECE 0.0177, mean probability on the true
Hotel B price 0.6926, and prior argmax accuracy 76.86\%. The selected run uses
entropy coefficient 0. A small-entropy robustness run (\(0.01\)) was nearly
identical: RevPAR gap \(+0.111\), occupancy gap \(+0.0029\), ADR gap \(-0.38\),
L1 0.0196, JS 0.0001. This makes the mechanism cleaner: stochasticity is not
being injected by a generic entropy bonus, but carried by the market prior and
KL-regularized policy.
We also tested an entropy-aware adaptive \(\beta\) rule. It showed no clear
practical gain over fixed \(\beta\), which is useful negative evidence: the
repair is not driven by elaborate coefficient tuning.

\section{Relation To Prior Methods And Scope}

Trace-Prior RL is close to behavior-regularized and KL-regularized RL. AWAC,
BRAC, MPO, Distral, KL-control, and Way Off-Policy all optimize a task
objective while penalizing drift from a reference behavior or policy
\citep{nair2020awac,wu2020brac,abdolmaleki2018mpo,teh2017distral,jaques2019way}.
CQL and BCQ are related conservative offline RL methods
\citep{kumar2020cql,fujimoto2019bcq}, but farther from our closed-loop online
setting. The reference in CP13 is not Hotel A's previous behavior and not
primarily an offline support constraint. It is an empirical estimate of the
market posterior predictive distribution under hidden competitor state.
Moreover, conservative offline RL primarily guards against out-of-distribution
actions; it does not by itself solve the Goodhart problem when the scalar reward
is the misspecified proxy.

The broader recipe is:
\begin{center}
\small learn trace prior + optimize reward inside a KL trust region.
\end{center}
This may apply when four ingredients coexist: the agent observes only a
projection of the true state, competent traces exist, the scalar reward is
easy to game, and the final system should improve rather than merely copy. In
LLM routing, for example, \(\piM(r\mid x)\) could be an expert route
distribution over models or tools, and the final policy could optimize
success-cost reward with KL discipline:
\[
\max_\theta
\mathbb{E}\left[
\mathrm{success}(x,r)-\alpha\,\mathrm{cost}(r)
-\beta R_{\mathrm{route}}(x)
\right],
\]
where
\[
R_{\mathrm{route}}(x)=
\KL(\pi_\theta(\cdot\mid x)\mid\mid\piM(\cdot\mid x)).
\]
This is only proposed relevance, not empirical generalization beyond hotels.
The current evidence is intentionally narrow: one controlled pricing simulator
with a fixed RM competitor.

A queueing-and-game-theory lens is a useful future direction rather than a
core proof burden for this paper. The two hotels can be viewed as competing
capacity-limited pipelines: guests route to Hotel A, Hotel B, or the outside
option, and price acts like a dynamic admission-control threshold. A low price
increases Hotel A's incoming flow but drains its finite inventory and changes
Hotel B's future load. This resembles a finite-horizon competing loss system or
dynamic pricing game more than a literal \(M/G/2\) queue, because guests do not
wait and hotel rooms are perishable capacity. Developing that connection could
turn the empirical failure mechanism into a higher-level theory of learning in
competitive service systems.

\paragraph{Limitations.}
First, Hotel B is fixed and deterministic. Real competitors may be noisy,
strategic, or learning. Second, \(\beta\) is reward-scale dependent. An
adaptive uncertainty-aware \(\beta\) is a natural extension, but our current
sensitivity runs showed no clear practical gain over fixed \(\beta\). Third,
Trace-Prior RL learns market behavior from traces; it does not recover Hotel
B's internal RM formula. Fourth, we do not yet include a second empirical
domain. The strongest claim is therefore a failure mechanism and repair recipe,
not broad empirical universality.

\section{Conclusion}

This case study shows why agent evaluation cannot stop at scalar reward. Hotel
A could look successful by RevPAR while failing to learn market-like yield
management. The trigger was partial observability: hidden competitor inventory
made the market label distributional, and deterministic decision rules
collapsed that uncertainty into shortcut behavior. Trace-level diagnostics
made the failure visible. Trace-Prior RL repaired it by learning the market
distribution from lagged traces and then optimizing Hotel A's own RevPAR inside
a KL trust region around that distribution. The result is a compact,
reproducible agent failure and fix: the score said success, the trace revealed
the wrong behavior, and the repair followed from the hidden-state mechanism.
The sharpest lesson is that, under hidden state, exact step-by-step prediction
accuracy and aggregate behavioral alignment can move in opposite directions.
The unit of contribution is therefore a disciplined agent-evaluation claim:
when an outcome score is too easy to satisfy, the trace decides whether the
agent has learned the intended behavior.

\bibliographystyle{plainnat}
\bibliography{references}

\end{document}